\documentclass[]{spie}  

\usepackage{amsmath,amsfonts,amssymb}
\usepackage{graphicx}
\usepackage[colorlinks=true, allcolors=blue]{hyperref}

\title{A Study on Improving Realism of Synthetic Data for Machine Learning}

\author[a]{Tingwei Shen}
\author[b]{Ganning Zhao}
\author[c]{Suya You}
\affil[a]{University of California, Berkeley, Berkeley, CA, USA}
\affil[b]{University of Southern California, Los Angeles, CA, USA}
\affil[c]{DEVCOM Army Research Laboratory, Los Angeles, CA, USA}

\authorinfo{Further author information: (Send correspondence to Tingwei Shen)\\Tingwei Shen: E-mail: tingweishen@berkeley.edu}

\begin{document} 
\maketitle

\begin{abstract}
Synthetic-to-real data translation using generative adversarial learning has achieved significant success in improving synthetic data. Yet, limited studies focus on deep evaluation and comparison of adversarial training on general-purpose synthetic data for machine learning. This work aims to train and evaluate a synthetic-to-real generative model that transforms the synthetic renderings into more realistic styles on general-purpose datasets conditioned with unlabeled real-world data. Extensive performance evaluation and comparison have been conducted through qualitative and quantitative metrics and a defined downstream perception task.
\end{abstract}


\section{INTRODUCTION}
\label{sec:intro}  

Deep learning has achieved remarkable success in recent years, driven partly by the availability of large amounts of labeled data. However, collecting and labeling data can be laborious and expensive, particularly in fields where data is scarce or difficult to obtain. Synthetic data offers a promising solution to this problem, as it can be generated automatically and at scale. However, synthetic data often lack the richness and diversity of real-world data, which limits its usefulness in many applications.

Thus, various studies have been proposed to refine synthetic data. Recent studies use contrastive learning (CL) for image-to-image translation\cite{10.1007/978-3-030-58545-7_19} and synthetic data refinement\cite{zhao2023unsupervised}. More studies have shown the success of using Generative adversarial networks (GANs) to improve the realism of synthetic data, as indicated by SimGAN\cite{shrivastava2017learning}, CycleGAN\cite{8237506}, DualGAN\cite{yi2018dualgan}, etc. GANs consist of two neural networks, a generator and a discriminator, which are trained with competing losses. The generator learns to produce synthetic data similar to the real data, while the discriminator learns to distinguish between the synthetic and real data. The generator gradually learns to produce more realistic data through this adversarial training.

In this paper, we aim to improve the realism of synthetic data for machine learning by training a synthetic-to-real generative model using GANs. Our approach builds on previous work in this area but focuses on evaluating and comparing different adversarial training setups on general-purpose datasets. We believe this will help identify the most effective techniques for improving the realism of synthetic data and ultimately lead to better performance in machine learning applications.

\section{Related Work}

Previous studies have explored using GANs to improve synthetic data for machine learning. For instance, Zhu et al. (2017)\cite{8237506} proposed the CycleGAN framework for unsupervised image-to-image translation, which can be used for style transfer and domain adaptation tasks. Shrivastava et al. (2017)\cite{shrivastava2017learning} proposed SimGAN, which uses a GAN to learn a mapping between simulated and real images, and the generated images are used to train a downstream task. SimGAN has been shown to improve the performance of object detection algorithms on synthetic data in a few-shot learning setting. Similarly, Atapattu\cite{8852449} proposed a GAN with the help of perceptual loss to make synthetic data more realistic. The perceptual loss measures the similarity of the synthetic images to the real images at a high-level semantic level, resulting in more realistic synthetic data.

Other studies have also explored using GANs for generating high-quality synthetic data. For instance, Zhang et al. (2019)\cite{zhang2019selfattention} proposed Self-Attention GAN (SAGAN), which improves the generation of high-resolution images. The SAGAN introduces a self-attention mechanism to capture long-range dependencies in the input data, resulting in more stable training and improved image quality. Moreover, Karras et al. (2019)\cite{karras2019stylebased} proposed StyleGAN, which enables the control of image attributes by learning disentangled representations of the image features. StyleGAN has been used for various applications, including face synthesis, image editing, and video synthesis. These studies and others demonstrate the potential of GANs for generating high-quality synthetic data and advancing various applications in computer vision and machine learning.

Despite these advances, there is still limited research on the effectiveness of adversarial training methods on general-purpose synthetic data for machine learning. This study aims to fill this gap by training and evaluating a synthetic-to-real generative model using GANs on general-purpose datasets, particularly self-driving scenes. The study will fine-tune an adversarial training setup and evaluate its performance through qualitative and quantitative metrics and downstream semantic segmentation tasks.

\section{Method}

In this section, we present the framework of our proposed model. Our model's structure is similar to a typical Generative Adversarial Network(GAN), which consists of a generator and a discriminator. The model consists of a refiner ($R_\theta(x)$) that generates refined images and a discriminator ($D_\phi$) that classifies images into synthetic and real. The model aims to learn a refiner that refines synthetic images ${x}_i$ to make it realistic to fool the discriminator. Our work is similar to SimGAN\cite{shrivastava2017learning}, where the inputs to our model are synthetic images. 

\subsection{Discriminator Model}
The discriminator, $D_\phi$, is trained to classify images as real vs refined, where $\phi$ are the parameters of the discriminator network. An input of the discriminator is a 3-channel RGB image, and the output is a binary number indicating whether the input image is synthetic or real. The training inputs are a set of refined images ${x}'$, where ${x}' = R_\theta(x)$, and a set of unlabeled real images ${y}_i$.  We update the parameters by minimizing the following loss:

\begin{equation}
    L_D(\phi) = - \sum_{i} {\log(D_\phi({x'}_i))} - \sum_{j} {\log(D_\phi({y}_j))}
\end{equation}

\subsection{Refiner Model}
The refiner, $R_\theta$, is trained to refine a synthetic image $x$ given a set of unlabeled real images $Y$. It takes in a synthetic image and outputs a refined image. The refiner should make the synthetic images look more realistic while preserving the annotation information of synthetic images. Thus, we define our refiner loss in two parts as the following:

\begin{equation}\label{refiner loss}
    L_R(\theta) = \alpha \sum_{i} {L_{Adv}(\theta; x_i; Y)} + \beta{L_{rec}(\theta; x_i)}
\end{equation}

\begin{equation}\label{adversarial loss}
    L_{Adv}(\theta; x_i; Y) = - \log(1 - D_\phi({x'}_i))
\end{equation}

\begin{equation}\label{self-regularization loss}
    L_{rec}(\theta; x_i; Y) = ||\Psi(R_\theta(x_i)) - \Psi(x_i)||
\end{equation}

\begin{equation}\label{perceptual loss}
    L_{perc}(\theta; x_i; Y) = \frac{1}{C_jH_jW_j} ||\psi_j(R_\theta(x_i)) - \psi_j(x_i)||^2
\end{equation}

$L_{Adv}(\theta; x_i; Y)$ is the adversarial loss that adds realism to the synthetic image $x_i$. It is defined by Equation \eqref{adversarial loss}, where we ask the discriminator to classify refined images and request it to fail to classify refined images as synthetic. 

$L_{rec}(\theta; x_i)$ is the reconstruction loss that preserves the annotation information of synthetic images. We use a self-regularization loss defined by Equation \eqref{self-regularization loss} during full training, where $\Psi$ is the mapping from image space to a feature space, and $||\cdot||$ is an L1 norm.

$\alpha$ and $\beta$ in Equation \eqref{refiner loss} are constants.

Note that we use a perceptual loss\eqref{perceptual loss} instead of self-regularization loss during pre-training of the refiner. Following Atapattu's work\cite{8852449}, the perceptual loss is defined by Equation \eqref{perceptual loss}, where $\psi$ is a pre-trained inception network, $\psi_j(x)$ is the activation of the $j$th convolution layer of the inception model and with a feature map of shape $C_j*H_j*W_j$. By minimizing the perceptual loss, we can make refined images perceptually similar to synthetic images in terms of high-level features such as object shapes and textures.

This paper uses Keras's inception v3 model as $\psi$ in the perceptual model and identity map as $\Psi$ in the self-regularization loss for implementation.

\subsection{Training  Algorithms}

Our goal is to train a refiner network $R_\theta$.
We first pre-train the refiner using perceptual loss with both inputs and supervision as synthetic images. In this way, the refiner learns an identity network. We also pre-train the discriminator with a set of synthetic and real images. 

Then we start the full training phase, where we alternatively update $\theta$ by taking an SGD step on $L_R(\theta)$ and update $\phi$ by taking an SGD step on $L_D(\phi)$. In practice, we update the refiner twice and then update the discriminator once in each training step. While updating the refiner, we keep the parameters of the discriminator fixed. While updating the discriminator, we keep the parameters of the refiner fixed, i.e., we use the current $\theta$ to compute $x'_i = R_\theta(x_i)$.

Similar to SimGAN\cite{shrivastava2017learning}, we also update the discriminator using a history of refined images during full training.

\section{Datasets}

\subsection{GTA-V to Cityscapes}
The Cityscapes dataset is a large-scale dataset consisting of high-quality images of real urban street scenes and pixel-level annotations of 30 different classes of objects. It has become a benchmark dataset for semantic segmentation, instance segmentation, and object detection tasks in computer vision. It contains over 5,000 images with a resolution of 2048x1024 pixels and has been widely used for research and development of machine learning models.

The GTA-V dataset is a computer vision dataset consisting of over 24,000 frames captured from the Grand Theft Auto V video game. The frames contain images of a simulated environment and corresponding pixel-level annotations of objects and their attributes.

There are 19 common semantic classes in Cityscapes and GTA-V dataset, so we will use GTA-V to Cityscapes as our synthetic-to-real dataset.

\section{Experiments}

\subsection{Implementation Details}
Due to computing limits, we downsample each synthetic and real image into (height, width) = (80, 160) by cropping and resizing. We start cropping at pixel row 200 and column 600, with a height of 400 pixels and width of 600 pixels, to get images with more objects of interest, and then resize the cropped image to (80, 160). 

For the refiner, we take $\alpha = 50$ and $\beta = 4*10^{-6}$ in Equation \eqref{refiner loss}. We pre-train the refiner for 1200 steps and the discriminator for 400 steps. The refiner learning rate is $lr = 0.0001$, and the discriminator learning rate is $lr = 0.001$. The architecture of our refiner network is based on residual network (ResNet)\cite{resNet}. An $80*160$ sized input image is convolved with $4*4$ filters that output 64 feature maps, and the output is passed through 5 ResNet blocks, which each consist of two convolutional layers with 64 feature maps and $4*4$ filters. The output of the last ResNet block is then passed into a $1*1$ convolutional layer to produce the refined image. The architecture of our discriminator network consists of 5 convolutional layers and 1 max pooling layer, which are shown in Table \ref{Dicriminator Architecture}.

During the full train phase, where we alternatively update the parameters of the refiner and the discriminator, we keep track of generator losses, discriminator losses for real images, and discriminator losses for refined images. Our target refiner is well-learned when the generator losses and discriminator losses for real images are the lowest, and the discriminator losses for refined images are the highest.

\begin{table}[ht]
\begin{center}       
\begin{tabular}{|c|c|c|c|c|}
\hline
\rule[-1ex]{0pt}{3.5ex}  Layer Number & Convolution Layer & Stride & Map Size & Activation  \\
\hline
\rule[-1ex]{0pt}{3.5ex}  1 & Conv $5 * 5$ & 3 & 96 & Relu \\
\hline
\rule[-1ex]{0pt}{3.5ex}  2 & Conv $4 * 4$ & 2 & 64 & Relu \\
\hline
\rule[-1ex]{0pt}{3.5ex}  3 & MaxPool $3 * 3$ & 2 & - & - \\
\hline
\rule[-1ex]{0pt}{3.5ex}  4 & Conv $3 * 3$ & 2 & 32 & Relu  \\
\hline
\rule[-1ex]{0pt}{3.5ex}  5 & Conv $1 * 1$ & 1 & 32 & Relu  \\
\hline
\rule[-1ex]{0pt}{3.5ex}  6 & Conv $1 * 1$ & 1 & 2 & Relu  \\
\hline
\rule[-1ex]{0pt}{3.5ex}  7 & SoftMax & - & - & -  \\
\hline
\end{tabular}
\end{center}
\caption{Architecture of the Discriminator Network} 
\label{Dicriminator Architecture}
\end{table}

\subsection{Evaluation Metrics}

\textbf{Frechet Inception Distance (FID)}, proposed by Martin Heusel\cite{FID}, is a metric to evaluate the quality of images created by generated models. It measures the distance between the distribution of generated results and the distribution of real images. A smaller FID score represents a closer distance between our refined results and real data renderings. 

\textbf{Structural Similarity Index(SSIM)} measures the similarities of two images. It is based on three comparison measurements -- luminance, contrast, and structure. In this work, we use Tensorflow's tf.image.ssim function to measure this metric.

\subsection{Downstream Segmentation Task Evaluations}

Except for evaluating our GAN model using metrics, we use downstream segmentation tasks to measure the quality of refined images quantitatively. We used two state-of-arts segmentation models: DeepLabv3+\cite{chen2018encoderdecoder} and U-NET\cite{ronneberger2015unet}. The inputs are 3-channel RGB images, and the outputs are indexed images, where each index indicates the class of the corresponding pixel. We evaluate the segmentation results using pixel accuracy and mIoU.

\subsection{Baseline and Comparison Model}
We use the synthetic-to-real dataset to train our proposed GAN model and get a set of refined images. Here, we have three image types: synthetic, refined, and real. The baseline is to use synthetic images to train the segmentation model and real images to test. We also use the same dataset to train the SimGAN\cite{shrivastava2017learning} model and get a set of SimGAN refined images. The segmentation model trained by SimGAN refined images is used for comparison.

\subsection{Results}

\begin{table}[ht]
\begin{center}       
\begin{tabular}{|l|l|l|}
\hline
\rule[-1ex]{0pt}{3.5ex}  Image Type & FID & SSIM  \\
\hline
\rule[-1ex]{0pt}{3.5ex}  Synthetic vs. Real (Baseline) & 60.82 & 0.0023   \\
\hline
\rule[-1ex]{0pt}{3.5ex}  Image after Pre-training Phase vs. Real & 56.89 & 0.0025  \\
\hline
\rule[-1ex]{0pt}{3.5ex}  Our Method Refined vs. Real & 55.71 & 0.0026  \\
\hline
\rule[-1ex]{0pt}{3.5ex}  SimGAN Refined vs. Real & 57.04 & 0.0025  \\
\hline
\end{tabular}
\end{center}
\caption{Evaluation Metrics of Synthetic/Refined Images vs. Real Images.} 
\label{FID, SSIM}
\end{table}

   \begin{figure} [ht]
   \begin{center}
   \begin{tabular}{c|c|c} 
   \includegraphics[height=3.5cm]{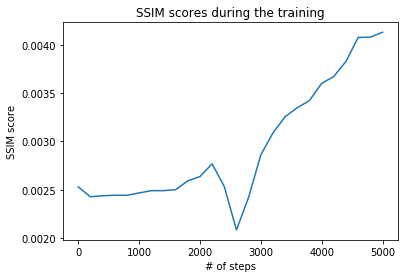} & \includegraphics[height=3.5cm]{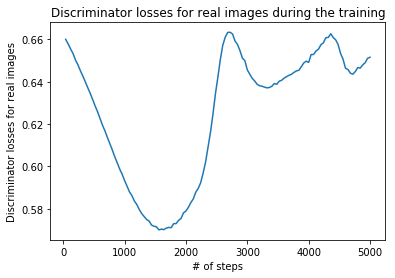} & \includegraphics[height=3.5cm]{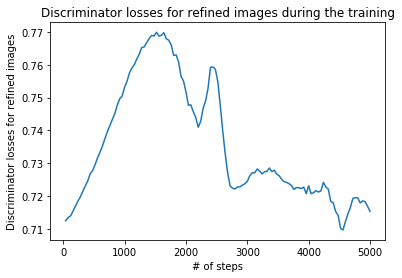} 
   \end{tabular}
   \end{center}
   \caption[example] 
   { \label{SSIM and overfit} 
Records from the GAN training in a total of 5000 steps. The left-hand image records SSIM scores during training. The middle image records the discriminator loss for real images. The right-hand image records the discriminator loss for refined images.}
   \end{figure}

\begin{table}[ht]
\begin{center}       
\begin{tabular}{|l|l|l|l|}
\hline
\rule[-1ex]{0pt}{3.5ex}  Training Image Type $\setminus$ Testing Image Type & Synthetic & Refined & Real  \\
\hline
\rule[-1ex]{0pt}{3.5ex}  Synthetic & 0.34 & 0.13 & 0.1 (Baseline)   \\
\hline
\rule[-1ex]{0pt}{3.5ex}  SimGAN Refined & 0.26 & 0.31 & 0.1  \\
\hline
\rule[-1ex]{0pt}{3.5ex}  Real & 0.14 & 0.11 & 0.49  \\
\hline
\hline
\rule[-1ex]{0pt}{3.5ex}  Our Method Refined & 0.23 & 0.43 & 0.14  \\
\hline
\end{tabular}
\end{center}
\caption{U-NET Downstream Segmentation Task's mIoU of Synthetic, Refined Images vs. Real Images
(Each element in the table is the mIoU of 200 test samples of the corresponding image type)} 
\label{segmentation mIoU}
\end{table}

\begin{table}[ht]
\begin{center}       
\begin{tabular}{|l|l|l|l|}
\hline
\rule[-1ex]{0pt}{3.5ex}  Training Image Type $\setminus$ Testing Image Type & Synthetic & Refined & Real  \\
\hline
\rule[-1ex]{0pt}{3.5ex}  Synthetic & 0.28 & 0.11 & 0.08 (Baseline)   \\
\hline
\rule[-1ex]{0pt}{3.5ex}  SimGAN Refined & 0.18 & 0.23 & 0.09  \\
\hline
\rule[-1ex]{0pt}{3.5ex}  Real & 0.12 & 0.1 & 0.42  \\
\hline
\hline
\rule[-1ex]{0pt}{3.5ex}  Our Method Refined & 0.15 & 0.31 & 0.12 \\
\hline
\end{tabular}
\end{center}
\caption{Deeplabv3+ Downstream Segmentation Task's mIoU of Synthetic, Refined Images vs. Real Images
(Each element in the table is the mIoU of 200 test samples of the corresponding image type)} 
\label{deeplab segmentation mIoU}
\end{table}

\begin{table}[ht]
\begin{center}       
\begin{tabular}{|l|l|l|l|}
\hline
\rule[-1ex]{0pt}{3.5ex}  Training Image Type $\setminus$ Testing Image Type & Synthetic & Refined & Real  \\
\hline
\rule[-1ex]{0pt}{3.5ex}  Synthetic & 78.55 & 31.97 & 32.8 (Baseline)   \\
\hline
\rule[-1ex]{0pt}{3.5ex}  SimGAN Refined & 74.34 & 77.3 & 33.33  \\
\hline
\rule[-1ex]{0pt}{3.5ex}  Real & 48.83 & 42.48 & 85.42  \\
\hline
\hline
\rule[-1ex]{0pt}{3.5ex}  Our Method Refined & 66.37 & 79.58 & 57.11  \\
\hline

\end{tabular}
\end{center}
\caption{U-NET Downstream Segmentation Task's Pixel Accuracy of Synthetic, Refined Images vs. Real Images (Each element in the table is the pixel accuracy in \% of 200 test samples of the corresponding image type)} 
\label{segmentation pixel accuracy}
\end{table}

\begin{table}[ht]
\begin{center}       
\begin{tabular}{|l|l|l|l|}
\hline
\rule[-1ex]{0pt}{3.5ex}  Training Image Type $\setminus$ Testing Image Type & Synthetic & Refined & Real  \\
\hline
\rule[-1ex]{0pt}{3.5ex}  Synthetic & 72.65 & 29.14 & 28.05 (Baseline)   \\
\hline
\rule[-1ex]{0pt}{3.5ex}  SimGAN Refined & 67.3 & 70.9 & 31.75 \\
\hline
\rule[-1ex]{0pt}{3.5ex}  Real & 46.39 & 39.38 & 82.9  \\
\hline
\hline
\rule[-1ex]{0pt}{3.5ex}  Our Method Refined & 63.5 & 72.79 & 56.43  \\
\hline

\end{tabular}
\end{center}
\caption{Deeplabv3+ Downstream Segmentation Task's Pixel Accuracy of Synthetic, Refined Images vs. Real Images (Each element in the table is the pixel accuracy in \% of 200 test samples of the corresponding image type)} 
\label{deeplab segmentation pixel accuracy}
\end{table}

   \begin{table} [ht]
   \begin{center}
   \begin{tabular}{l|c|c} 
    & Sample Images & Sample Results of the Segmentation Network \\
    & & Trained on Corresponding Image Types and Test on Real Images \\ 
    \hline
    Synthetic & \begin{tabular}{@{}c@{}} \\ \includegraphics[height=1.8cm]{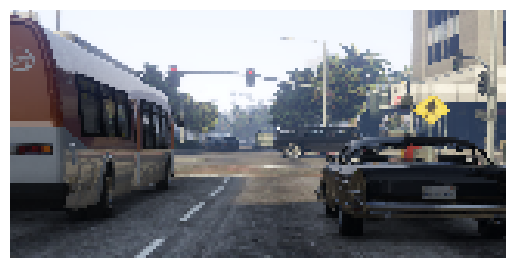}  \\ \includegraphics[height=1.8cm]{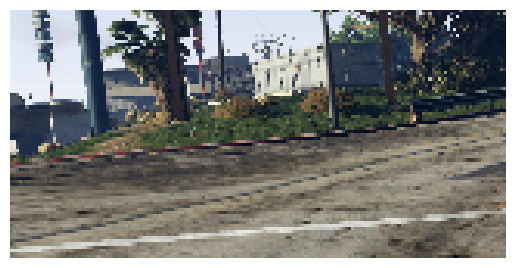} \\ \includegraphics[height=1.8cm]{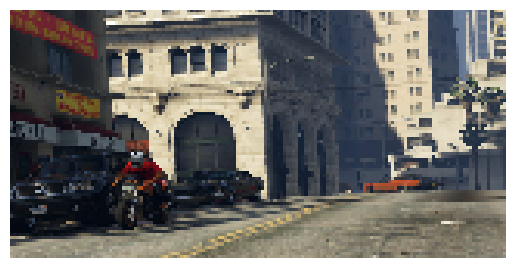} \\ \end{tabular}   & \begin{tabular}{@{}c@{}} \\ \includegraphics[height=1.8cm]{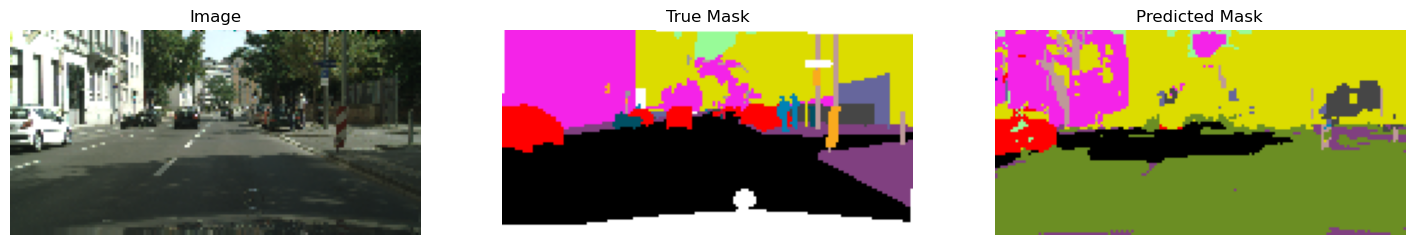} \\ \includegraphics[height=1.8cm]{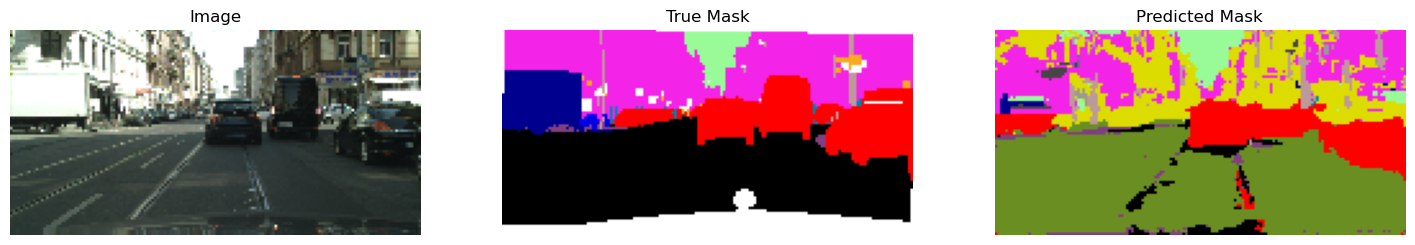} \\ \includegraphics[height=1.8cm]{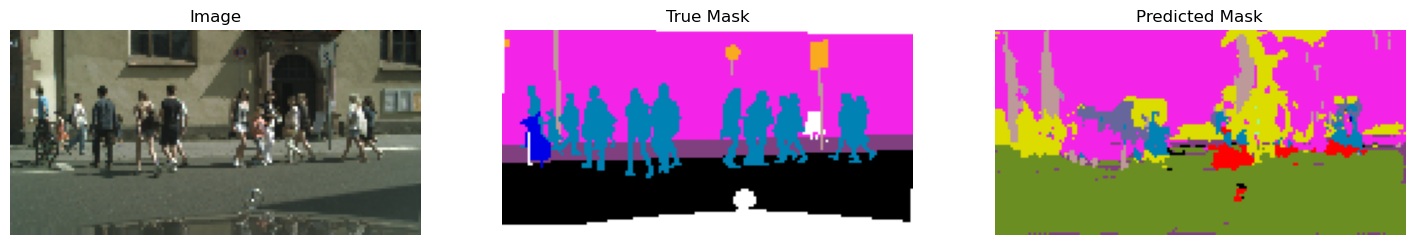}
   \end{tabular}\\
    \hline
   SimGAN & \begin{tabular}{@{}c@{}} \\ \includegraphics[height=1.8cm]{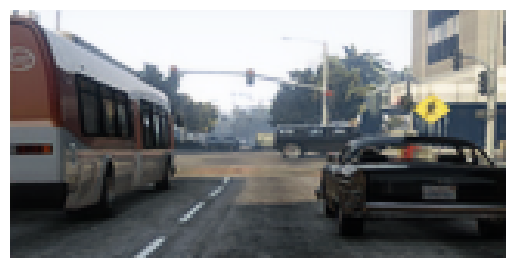}  \\ \includegraphics[height=1.8cm]{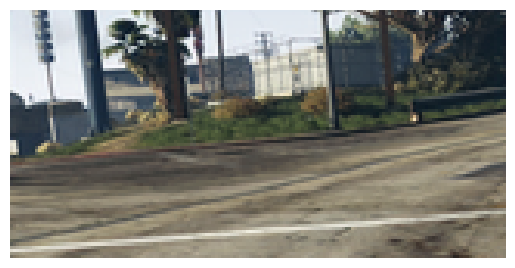} \\ \includegraphics[height=1.8cm]{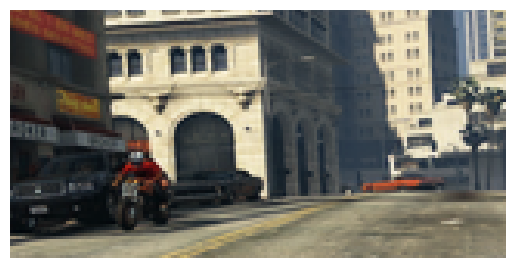} \\ \end{tabular} & \begin{tabular}{@{}c@{}} \\ \includegraphics[height=1.8cm]{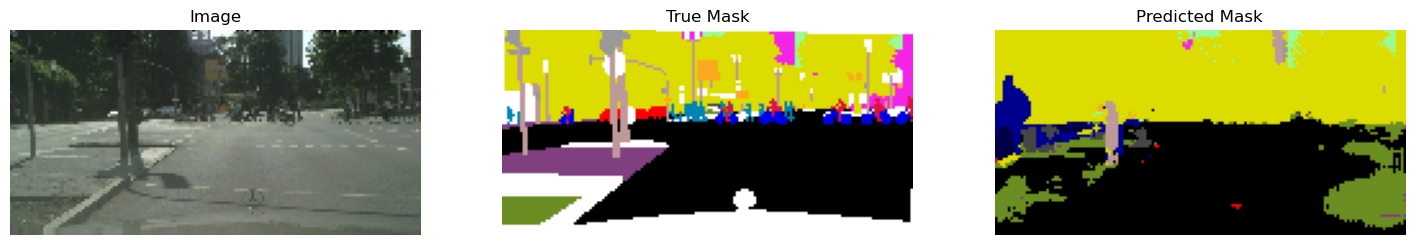} \\ \includegraphics[height=1.8cm]{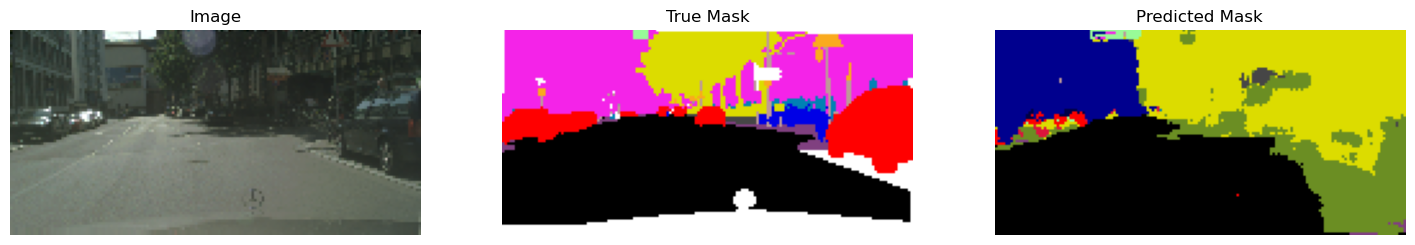} \\ \includegraphics[height=1.8cm]{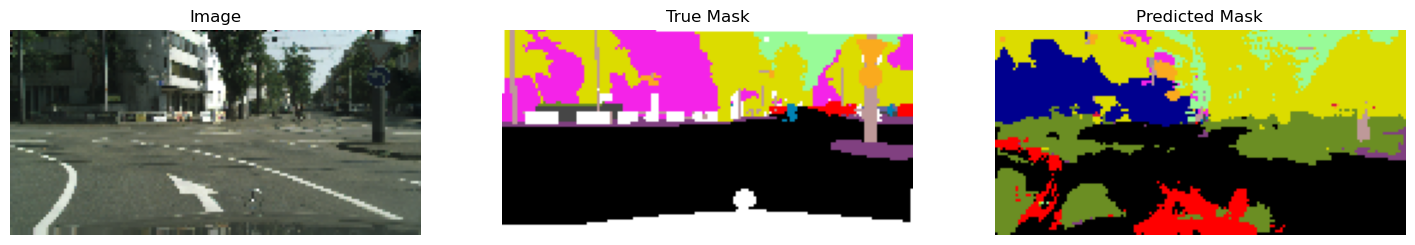}
   \end{tabular} \\
   \hline
   Our Method & \begin{tabular}{@{}c@{}} \\ \includegraphics[height=1.8cm]{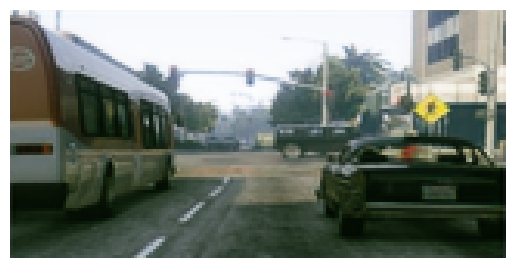}  \\ \includegraphics[height=1.8cm]{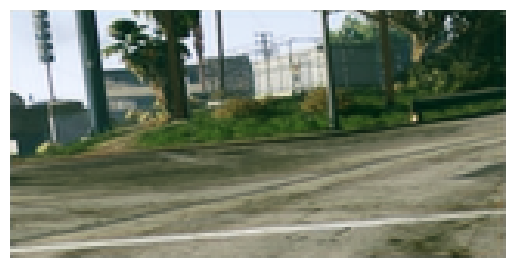} \\ \includegraphics[height=1.8cm]{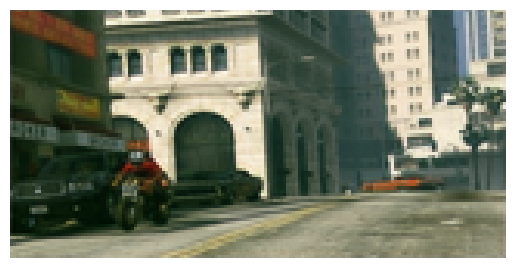} \\ \end{tabular} & 
   \begin{tabular}{@{}c@{}} \\ \includegraphics[height=1.8cm]{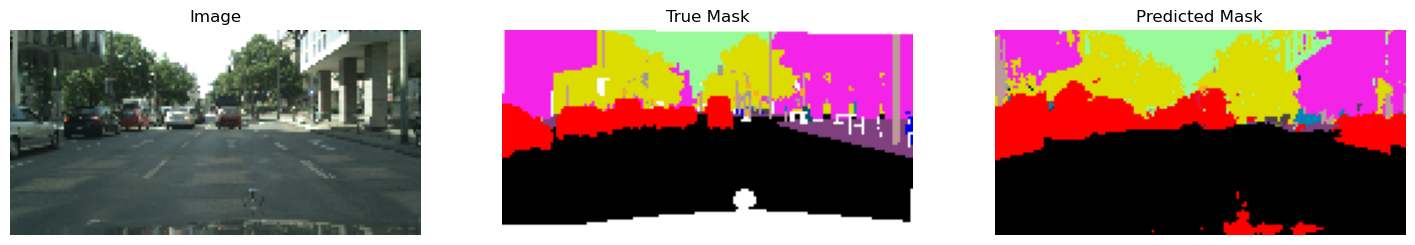} \\ \includegraphics[height=1.8cm]{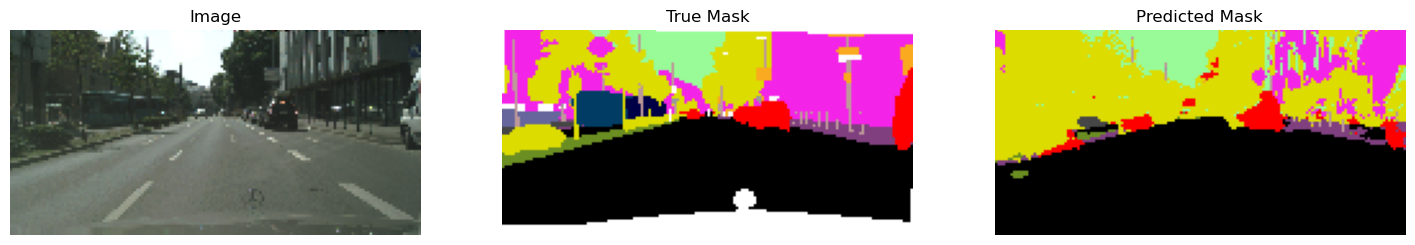} \\ \includegraphics[height=1.8cm]{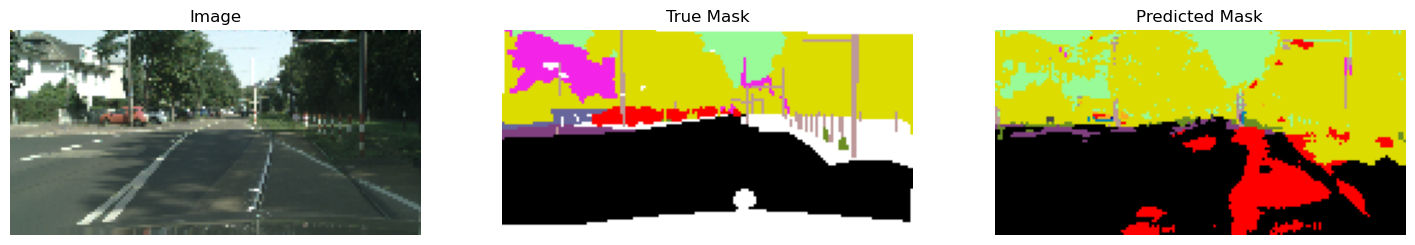}
   \end{tabular}  \end{tabular}
   \end{center}
   \caption[example] 
   { \label{Visual Segmentation Results} 
Visual Results of Downstream Segmentation Task. (Image type from left to right: sample synthetic/refined image for training, sample real image for testing, the ground truth of the real image, the predicted mask of the real image)}
   \end{table}

With the implementation details explained in 5.1, we get a trained refiner network that outputs refined images. We sample synthetic and real images and calculate their FID and SSIM scores as the baseline. The results are shown in Table \ref{FID, SSIM}, where we see a lower FID score and a higher SSIM score for refined images, which means the refined images have a closer distribution to the real ones. Compared with the images after pre-training but before the full training phase and with the SimGAN refined images, our refined images are also closer to the real renderings.

If we plot out the FID and SSIM scores of refined images versus real images during training, we find their trends consistent, i.e., when the FID score goes down, the SSIM score goes up. Yet a lower FID score and a higher SSIM score don't mean a good-quality refiner network or a better dataset for downstream segmentation tasks. Referring to Figure \ref{SSIM and overfit}, the SSIM score generally increases as we train more steps. Yet, the optimal step for a refiner is at 1800, where the discriminator loss for real images is the lowest and for refined images is the highest, indicating that the discriminator successfully learns to classify real images as real. The refiner successfully fools the discriminator into classifying more refined images as real. As we pass 1800 steps, the refiner overfits, so the refined images are not good for training segmentation networks even though they have a closer distribution to the real images, as indicated by SSIM scores.

Given a trained refiner, we get three different image types: synthetic, refined, and real images. We use each pair to train and test a downstream segmentation task, using U-NET and Deeplabv3+ segmentation models, and the results are indicated in Table \ref{segmentation mIoU}, \ref{deeplab segmentation mIoU}, \ref{segmentation pixel accuracy} and \ref{deeplab segmentation pixel accuracy}. The first column indicates the training data type, and the first row indicates the testing data type. The last row is trained by our refined data. Overall, the U-NET model fits better with our data, but both models show a consistent relationship between synthetic, refined, and real data. The diagonal terms use the same training and testing dataset, so they are the highest per column and row. Particularly, models trained and tested by the real dataset have the best accuracy and mIoU as expected, which drove our purpose of mimicking the real dataset. Observing that the results of models trained and tested by our method refined data(last row, third column) is better than those trained and tested by SimGAN method refined data(third row, third column) and by synthetic data(second row, second column), we found our refined data forms a better dataset for training segmentation models. Compared with the baseline, the segmentation network trained by refined images significantly improves testing real datasets, indicating that our refined images are more realistic and form a better dataset in training segmentation tasks on real-world data. Although SimGAN refined data also shows some improvement of segmenting real data, it is less helpful than our refined data. Observing the third column when the refined images are used for testing, the values are generally low. Our refined images behave better as a training dataset than a testing dataset.

 We present our visual results in Table \ref{Visual Segmentation Results}. The segmentation network trained on our refined images could classify pixels of real images more accurately than the network trained on synthetic or SimGAN refined images. Visually, our refined images capture a closer color tone to the cityscapes dataset while keeping the contents of the original synthetic images unchanged. SimGAN refined images are more similar to synthetic images and are more blurry. This improvement is largely due to our newly induced perceptual loss during the pre-training phase.

\section{Conclusion}

In conclusion, the results obtained from implementing the refiner network and downstream segmentation tasks show that the refined images generated by the network have a closer distribution to real images, as indicated by the lower FID score and higher SSIM score. The segmentation network trained using these refined images performs significantly better in testing real datasets than the baseline, indicating that the refined images form a better dataset for training segmentation tasks on real-world data. However, our refined images behave worse as a testing dataset than a training dataset. The comparison with SimGAN also shows that the segmentation network trained on refined images outperforms those trained on synthetic or SimGAN-refined images, both visually and quantitatively. The perceptual loss induced in the pre-training phase better reconstructs the image. Therefore, our refiner network can effectively improve the realism of synthetic images and provide a better dataset for downstream segmentation tasks.

However, there are some limitations to our study. The image size we used for training is small due to computing limits, so we lose a lot of information from original images. Future directions include using more synthetic-to-real datasets, training with larger image sizes, and exploring more efficient methods.

\bibliography{report} 

\begin{thebibliography}{10}

\bibitem{10.1007/978-3-030-58545-7_19}
Park, T., Efros, A.~A., Zhang, R., and Zhu, J.-Y., ``Contrastive learning for
  unpaired image-to-image translation,'' in [{\em Computer Vision – ECCV
  2020: 16th European Conference, Glasgow, UK, August 23–28, 2020,
  Proceedings, Part IX}{\nolinebreak\hspace{0.1em}]},   319–345,
  Springer-Verlag, Berlin, Heidelberg (2020).

\bibitem{zhao2023unsupervised}
Zhao, G., Shen, T., You, S., and Kuo, C. C.~J., ``Unsupervised synthetic image
  refinement via contrastive learning and consistent semantic and structure
  constraints,'' (2023).

\bibitem{shrivastava2017learning}
Shrivastava, A., Pfister, T., Tuzel, O., Susskind, J., Wang, W., and Webb, R.,
  ``Learning from simulated and unsupervised images through adversarial
  training,'' (2017).

\bibitem{8237506}
Zhu, J.-Y., Park, T., Isola, P., and Efros, A.~A., ``Unpaired image-to-image
  translation using cycle-consistent adversarial networks,'' in [{\em 2017 IEEE
  International Conference on Computer Vision
  (ICCV)}{\nolinebreak\hspace{0.1em}]},   2242--2251 (2017).

\bibitem{yi2018dualgan}
Yi, Z., Zhang, H., Tan, P., and Gong, M., ``Dualgan: Unsupervised dual learning
  for image-to-image translation,'' (2018).

\bibitem{8852449}
Atapattu, C. and Rekabdar, B., ``Improving the realism of synthetic images
  through a combination of adversarial and perceptual losses,'' in [{\em 2019
  International Joint Conference on Neural Networks
  (IJCNN)}{\nolinebreak\hspace{0.1em}]},   1--7 (2019).

\bibitem{zhang2019selfattention}
Zhang, H., Goodfellow, I., Metaxas, D., and Odena, A., ``Self-attention
  generative adversarial networks,'' (2019).

\bibitem{karras2019stylebased}
Karras, T., Laine, S., and Aila, T., ``A style-based generator architecture for
  generative adversarial networks,'' (2019).

\bibitem{resNet}
He, K., Zhang, X., Ren, S., and Sun, J., ``Deep residual learning for image
  recognition,'' (2015).

\bibitem{FID}
Heusel, M., Ramsauer, H., Unterthiner, T., Nessler, B., and Hochreiter, S.,
  ``Gans trained by a two time-scale update rule converge to a local nash
  equilibrium,'' in [{\em Advances in Neural Information Processing
  Systems}{\nolinebreak\hspace{0.1em}]},  Guyon, I., Luxburg, U.~V., Bengio,
  S., Wallach, H., Fergus, R., Vishwanathan, S., and Garnett, R., eds.,  {\bf
  30}, Curran Associates, Inc. (2017).

\bibitem{chen2018encoderdecoder}
Chen, L.-C., Zhu, Y., Papandreou, G., Schroff, F., and Adam, H.,
  ``Encoder-decoder with atrous separable convolution for semantic image
  segmentation,'' (2018).

\bibitem{ronneberger2015unet}
Ronneberger, O., Fischer, P., and Brox, T., ``U-net: Convolutional networks for
  biomedical image segmentation,'' (2015).

\end{thebibliography}
\bibliographystyle{spiebib} 

\end{document}